# An Effective Training Method For Deep Convolutional Neural Network

Yang Jiang[1*], Zeyang Dou[1*], Qun Hao[1,2†], Jie Cao[1], Kun Gao[1], Xi Chen[3]

1. School of Optoelectronic, Beijing Institute of Technology. Beijing, China, 100081
2. Graduate School at Shenzhen, Tsinghua University, Tsinghua Campus, The University Town, Shenzhen, China, 518055.
3. BGI Research, Beishan Industrial Zone, Yantian District, Shenzhen, China, 518083
† Correspondence: qhao@bit.edu.cn

**Abstract**

In this paper, we propose the nonlinearity generation method to speed up and stabilize the training of deep convolutional neural networks. The proposed method modifies a family of activation functions as nonlinearity generators (NGs). NGs make the activation functions linear symmetric for their inputs to lower model capacity, and automatically introduce nonlinearity to enhance the capacity of the model during training. The proposed method can be considered an unusual form of regularization: the model parameters are obtained by training a relatively low-capacity model, that is relatively easy to optimize at the beginning, with only a few iterations, and these parameters are reused for the initialization of a higher-capacity model. We derive the upper and lower bounds of variance of the weight variation, and show that the initial symmetric structure of NGs helps stabilize training. We evaluate the proposed method on different frameworks of convolutional neural networks over two object recognition benchmark tasks (CIFAR-10 and CIFAR-100). Experimental results showed that the proposed method allows us to (1) speed up the convergence of training, (2) allow for less careful weight initialization, (3) improve or at least maintain the performance of the model at negligible extra computational cost, and (4) easily train a very deep model.

## Introduction

Convolutional neural networks (CNNs) have enabled the rapid development of a variety of applications, particularly ones related to visual recognition tasks [He et al. 2015, 2016]. There are two major reasons for this: the building of more powerful models, and the development of more efficient and robust strategies. On the one hand, recent deep learning models are becoming increasingly complex owing to their increasing depth [He et al. 2016, Simonyan 2014, Szegedy 2015] and width [Zeiler 2014, Sermanet et al. 2013], and decreasing strides [Sermanet et al. 2013]; on the other hand, better generalization performance is obtained by using various regularization techniques [Ioffe 2015, Ba and Salimans 2016], designing models of varying structure [He et al. 2016, Huang et al. 2016] and new nonlinear activation functions [Goodfellow et al. 2013, Agostinelli et al. 2014, Eisenach et al. 2016].

However, most of the aforementioned techniques follow the same underlying hypothesis: the model is highly non-linear because many inputs directly fall into the nonlinear parts of the activation functions. Although the highly nonlinear structure improves model capacity, it leads to difficulties in training. Training problems have recently been partly addressed through carefully constructed initializations [Mishkin 2015, Salimans 2016] and batch normalization (BN) [Ioffe 2015]. However, these methods may lead to a loss of efficiency in the training of very deep neural networks. Even though ResNets outperform the plain CNNs, they still require a warming up trick to train a very deep network [He 2016]. Thus the challenges to training have not been completely solved.

As a counterpart, deep neural networks with linear acti-vation functions, which have relatively low model capacity, are relatively easy to optimize at the beginning of training, as we will show. Thus, we need to strike and maintain a balance between the difficulties in training and model ca-pacity. With regard to the possibility of combining the ad-vantages of low-capacity and high-capacity models, one motivating example is the multigrid method [Bakhvalov 1966] used to accelerate the convergence of differential equations using a hierarchy of discretization. The main idea underlying the multigrid method is to speed up computation by using a coarse grid and interpolating a correction computed from this into a fine grid. The model with low capacity can be considered a coarse approximation of the problem, and the high-capacity model corresponds to a fine approximation. Similarly, unsupervised pre-training [Hinton 2006] and pre-training with shallow networks [Simonyan 2014] correspond to the use of models with relatively low capacity to coarsely fit the training data. The pre-trained parameters are then transferred to the complex model to carefully fit the dataset.

---

*Authors contributed equally

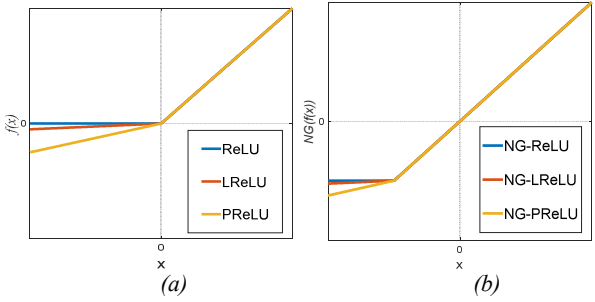 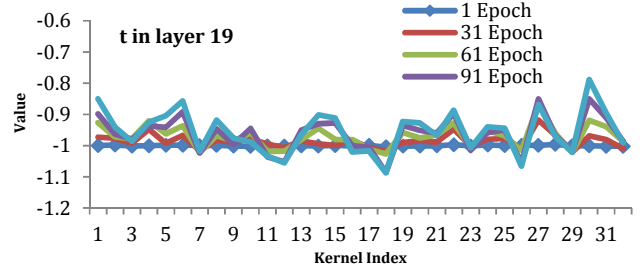

*Figure. 1 Examples of ReLU, LReLU and PReLU and their NG versions. (a) ReLU, LReLU and PReLU; (b) NG-ReLU, NG-LReLU and NG-PReLU, $t = -1$.*

*Fig. 2 The values of $t$ in layer 19 during the training procedure. The model is 56-layer plain CNN.*

Above all, it makes sense to combine the advantages of models of different capacity by first restricting capacity to coarsely fit the problem, and then endowing the model with greater capacity during the training procedure, hence enabling it to gradually perform better. Note that the symmetric structures of activation functions also make the training procedure stable, as we will show. Therefore we modify a family of activation functions such as Recified Linear Unit (ReLU), Leaky ReLU (LReLU) and Parametric ReLU (PReLU) by introducing a trainable parameter $t$, which we call nonlinearity generator (NG), to make the activation functions linear symmetric for the inputs at initial stage. NGs then introduce nonlinearity during the training procedure, endowing the model with greater capacity. The introduced parameter can be easily incorporated into the back-propagation framework. The proposed method allows us to (1) speed up the convergence of training, (2) allow for less careful parameter initializations, (3) improve or at least maintain the performance of convolutional neural network at negligible extra computational cost, and (4) easily train a very deep model.

## Approach

**Nonlinearity Generator**

We define the nonlinearity generator (NG) as follows

$$NG(f(x_i)) = f(x_i - t_i) + t_i \qquad (1)$$

Where $f$ is an activation function such as ReLU, LReLU and PReLU, $x_i$ is the input of activation function on the $i$ th node, and $t_i$ is a trainable parameter controlling the linearity of the generator given the input distribution. Note that $t_i$ is different for each node, we call equation (1) the element-wise version. We also consider a channel-wise variant: the different nodes in the same channel share the same $t$. ***If $t$ increases, we say that NG introduces greater nonlinearity because this increases the probability that inputs of the activation functions fall into the nonlinear parts.***

If $t$ is smaller than the minimum input, all inputs of the NG are in the linear area, making it a linear symmetric activation function for the inputs. Figure 1 compares the shapes of different activation functions (ReLU, LReLU and PReLU) and their NG versions (NG-ReLU, NG-LReLU and NG-PReLU). Given a preprocessed input image and some proper weight initialization, the property of NG can guarantee that the model is almost linear at the initial stage. As will show in analysis, the capacity of this model is relatively low, making training relatively easy. Thus, difficulties in training are alleviated during the initial iterations.

$t_i$ can be easily optimized using back propagation algorithm. Consider the element-wise version for example; the derivative of $\{t_i\}$ is simply followed the chain rule

$$\frac{\partial \varepsilon}{\partial t_i} = \frac{\partial \varepsilon}{\partial f(x_i)} \frac{\partial f(x_i)}{\partial t_i} \qquad (2)$$

where $\varepsilon$ represents the objective function, and

$$\frac{\partial f(x_i)}{\partial t_i} = \begin{cases} 0 & x_i > t_i \\ 1 & x_i \leq t_i \end{cases}. \quad (3)$$

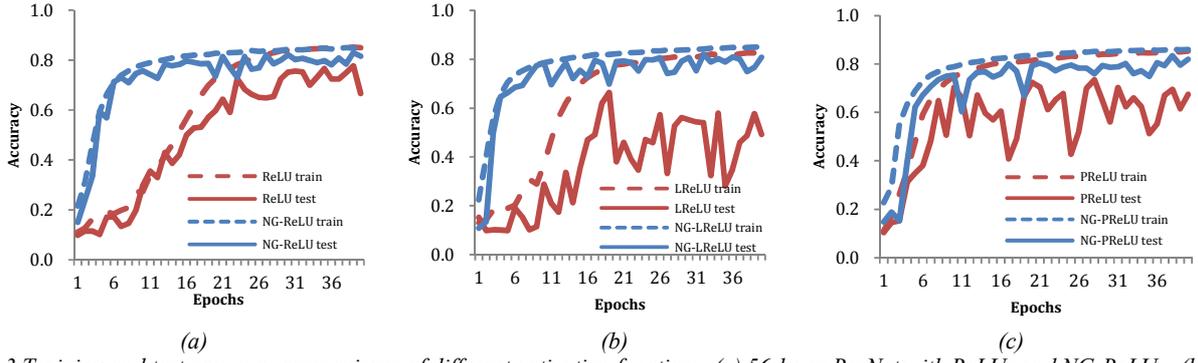

*Fig. 3 Training and test accuracy comparisons of different activation functions. (a) 56-layer ResNet with ReLUs and NG-ReLUs; (b) 56-layer ResNet with LReLUs and NG-LReLUs. (c) 56-layer ResNet with PReLUs and NG-PReLUs.*

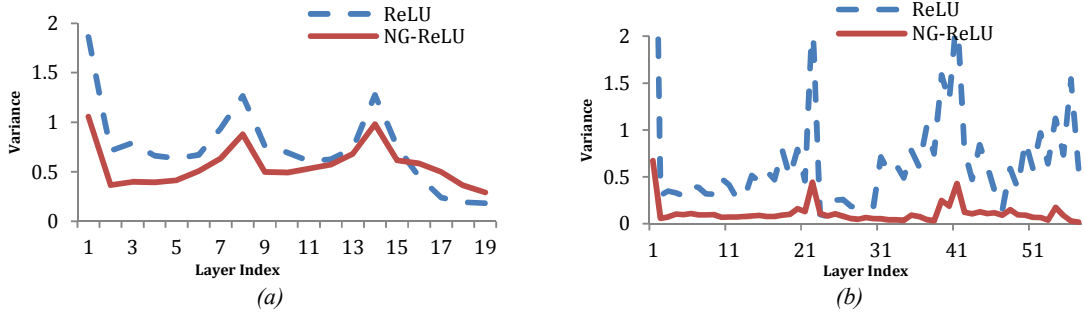

*Fig. 4 Weight variance comparisons of every layer. (a) 20- Layer plain CNN; (6) 56-layer ResNet.*

By using gradient descent, the NG can itself determine the degree of nonlinearity for each layer based on gradient information during the training process, endowing the model with greater capacity.

Since the experiments have shown that the performances of the channel-wise and element-wise versions are comparable, we use the former because it introduces very small number of extra parameters.

We adopt the momentum method when updating parameter $t$

$$\Delta t_i = \eta \Delta t_i + \gamma \frac{\partial \varepsilon}{\partial t_i}. \quad (4)$$

## Analysis

### NG Acts as a Regularizer

First, we use a toy example to show that NG with strong nonlinearity improves the capacities of models. We used a simple CNN without activation functions as a baseline to fit the test dataset CIFAR-10. The model had two convolutional layers, one dense layer without activation functions and a softmax layer. Each convolutional layer contained three $3 \times 3$ kernels, and the loss function was cross entropy. We used the stochastic gradient descend (SGD) optimizer to train the model. The initial learning rate was 0.01, and we divided it by 10 when the loss no longer decreased in value. The batch size was 128, and we adopt the momentum of 0.9, the same simple data augmentation as [He et al. 2016], and MSRA weight initialization [He et al. 2015].We then added NG-ReLUs with different untrainable values of $t$ for the two convolutional layers and make comparisons with the baseline model. Finaly, we initially set $t$ to -1 and made it trainable to test training performance. Table 1 shows the maximum training accuracies of the different models, where None means the baseline model without activation functions. We see that as $t$ increased, so did the maximum training accuracy. For the NG with trainable $t$, the mean value increased to -0.4765 at the end of training, and thus the performance was between that of -0.5 and -0.25 . This toy example shows that the nonlinearity of NG changes model capacity.

Table 1 Training accuracy comparisons

| $t$ | None | -2 | -1 | -0.5 | -0.25 | Trainable |
|---|---|---|---|---|---|---|
| Training Accuracy (%) | 35.75 | 35.75 | 42.66 | 46.20 | 50.03 | 46.73 |

Next, we experimentally show that CNN with less nonlinear NG is relatively easy to train. Our goal is to explore the critical depth, which is the depth beyond which the model does not converge. The test models were the plain CNNs with NG-ReLUs, the different parameters $t$ of which were untrainable. We then set $t = -1$ initially, made it as a trainable parameter and tested the critical depth. The model structure was the same as [He et al. 2016] without BN, and the dataset was CIFAR-10. We used the same training strategy except that Xavier initialization [Glorort 2010] were used as the weight initialization for all the experiments. Table 2 shows the results.

Table 2 Critical depth of CNN for different $t$

| $t$ | -1 | -0.75 | -0.5 | -0.25 | Trainable |
|---|---|---|---|---|---|
| Critical Depth | 116 | 116 | 104 | 92 | 116 |

We see that as $t$ increased, the critical depth decreased, indicating the increasing difficulty in training. As discussed previously, a greater value of $t$ corresponds to higher model capacity, and thus models with lower capacities are relatively easy to train.

Recall the properties of NG, it initially makes the model almost linear, and automatically enhance the model capacity during the training. Thus, NG can be considered an unusual form of regularization during the training procedure: parameters in the early stages are obtained by training a relatively low-capacity model for only a few iterations, and these parameters are reused for the initialization of a higher-capacity model. As all the inputs of NG were in the linear area using a proper value of $t$, the model was almost linear in the initial stage; as the training contained, $t$ was updated based on gradient information, hence endowing the model with greater capacity. We extracted $t$ from the 56-layer plain CNN to see how $t$ changed during the training procedure. Figure 2 shows the value of $t$ in layer 19 for different epochs. We see that as the training proceeded, $t$ became increasingly oscillatory, increasing the capacity of the model.

An advantage of this strategy is that the model is less likely to overfit in the early stage of the training procedure because we restrict the parameters to particular regions at first, following which they are expanded gradually to seek better parameters. As seen in Figure 3, although the training accuracies of ResNets with ReLUs, LReLUs and PReLUs increased steadily, their test accuracy oscillated. By comparison, the test accuracies of ResNets with NG-ReLUs, NG-LReLUs and NG-PReLUs were far more stable.

**Symmetric Structure of Activation Function Affects Training procedure**

Many researchers have shown that the approximately symmetric structure of activation functions can speed up learning because it alleviates the mean shift problem. Mean shift correction pushes the off-diagonal block of Fisher Information Matrix close to zero [He et al. 2015], rendering the SGD solver closer to the natural gradient descent [Clevert 2015]. In this subsection, we further show that variance of weight variation may influence the training as well. We have the following theorem.

**Theorem 1**: Given a fully-connected neural network with $M$ layers, let $W^l$ and $Z^l$ be the weight and the input of layer $l$, $S^l = W^l Z^l$, then variance of the weight variation has the following lower and upper bounds

$$\eta^2 E^2(\frac{\partial \varepsilon}{\partial S^l}) Var(Z^l) \leq Var(\Delta W^l) \leq 2\eta^2 (Var(\frac{\partial \varepsilon}{\partial S^l})E^2(Z^l) + Var(Z^l)Var(\frac{\partial \varepsilon}{\partial S^l}) + Var(Z^l)E^2(\frac{\partial \varepsilon}{\partial S^l})) \quad (5)$$

where $\eta$, $\Delta W^l$ and $\varepsilon$ represent the learning rate, the weight variation in layer $l$ and the objective function respectively.

Proof: Please see the appendix.

This theorem shows the variance of weight variation in layer $l$ is closely related to the expectation of the activations $E(Z^l)$ and gradients $E(\partial \varepsilon / \partial S^l)$. The asymmetric structure of the activation function may cause mean shift problem [Clevert 2015]. Mean shift caused by previous layer acts as bias for the next layer. The more the units are correlated, the higher their mean shift [Clevert 2015]. The shift of activation expectations in turn may affect the expectation of the gradient. From theorem 1, the mean shift problem makes the upper and lower bounds of variance $Var(\Delta W)$ in different layers unstable, thus they raise the instability of the variation of the weights in different layers. The unstable variation of the weights would result in unstable weight variance during the training, which hampers information flow [Glorot 2010], raising the training problems when we train a very deep model. To stabilize the training procedure, we want to keep the two bounds stable. Because the proposed NG is symmetric with respect to the original point for the linear area, it can pull $E(Z^l)$ and $E(\partial \varepsilon / \partial S^l)$ close to zero, making the training procedure more stable. Figure 4 shows the comparisons of weight variance for the 20-layer plain CNNs and 56-layer ResNets. Compared with ReLU, the weight variance using NG-ReLU was more stable, which supports our analysis above.

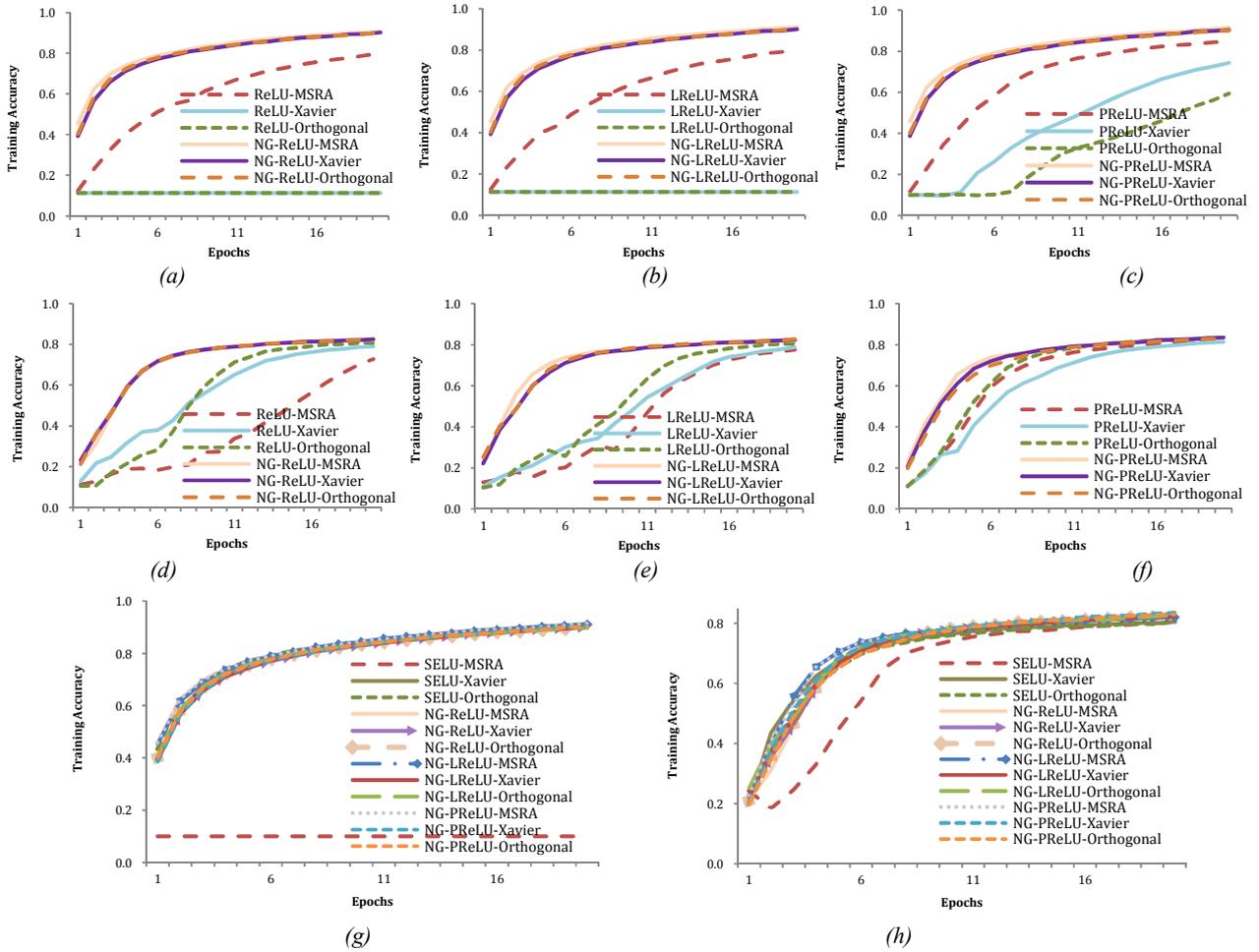

*Figure 5. Learning behavior comparisons with different initializations on CIFAR-10. (a), (b), (c), (g) plain CNNs; (d), (e), (f), (h) ResNets.*

## Experiments

We tested our method on two models (plain CNN without BN and ResNet, whose structures were the same as [He et al. 2016] and reported in the appendix). Our GPU was a GTX1080-ti. We focused on a fair performance comparisons of the models, not on state-of-the-art results. Thus, we used the same strategies during training. We modified the activation functions of the model, i.e., ReLU, LReLU, and PReLU, as NG-ReLU, NG-LReLU, and NG-PReLU, respectively, and compare the model performances with their counterparts and the scaled exponential linear units (SELU). The datasets are CIFAR-10 (color images in 10 classes, 50k train and 10k test) and CIFAR-100 (color images in 100 classes, 50k train and 10k test). We used the same simple data augmentation method as [He et al. 2016] for CIFAR-10 and CIFAR-100. $t$ were set to -1 for all experiments. For plain CNNs, the initial learning rate was 0.01, and we divided it by 10 during the training procedure when the test error no longer decreased. For ResNets, the initial learning rate was 0.1, and we divided it by 10 at 80, 120, and 160 epochs, and terminated training at 200 epochs. For both models, we used a batch size of 128 and a weight decay of 0.0005, and a momentum of 0.9. Note that we did not use other techniques, such as dropout [Srivastava et al. 2014], to reduce the effect of the random factors on the experimental results to a minimum.

**Learning Behavior**

We evaluated our method on the plain CNNs and ResNets over CIFAR-10. The test models were 20-layer plain CNNs and 56-layer ResNets with different activation functions, i.e. ReLUs, LReLUs, PReLUs, their NG versions (NG-ReLUs, NG-LReLUs, NG-PReLUs) and SELUs. Three parameter initialization methods—Xavier initialization, MSRA initialization, and orthogonal initialization [Saxe et al. 2013]—were used to test the behavior of the models. Fig. 5 shows the training accuracy comparisons of different models. We see the models with our method were insensitive to the different initializations, while the other models were not. In addition, the models with NGs converged much faster than their counterparts in all cases. For ResNets, our methods were still robust to the initializations and converged much faster than their counterparts, while the other models were relatively sensitive to the initializations.

**Model Performance**

We tested and compared various models, including 44-layer, 56-layer plain CNNs and ResNets, on CIFAR-10 and CIFAR-100 datasets. MSRA initialization was used. Note that plain CNNs with ReLUs, LReLUs, PReLUs and SELUs did not converge when the depths were more than 20 layers, whereas models with NGs still converged. Table 3 shows the test accuracy comparisons. Although our method first trained the models with the relatively low capacity, they still resulted better or at least comparable performances compared with their counterparts and SELUs in most cases.

Table 3 Test accuracy comparisons on different datasets. "-" means not convergent

| Model | Dataset | ReLU | LReLU | PReLU | SELU | NG-ReLU | NG-LReLU | NG-PReLU |
|---|---|---|---|---|---|---|---|---|
| Plain-44 | CIFAR-10 | - | - | - | - | **89.58%** | **88.44%** | **89.03%** |
| Plain-56 | CIFAR-10 | - | - | - | - | **88.72%** | **87.56%** | **87.95%** |
| ResNet-44 | CIFAR-10 | 92.83% | 92.71% | 93.23% | 92.66% | **93.48%** | **93.18%** | **93.33%** |
| ResNet-56 | CIFAR-10 | 92.45% | 92.18% | 93.28% | 92.57% | **93.90%** | **93.56%** | **93.49%** |
| Plain-44 | CIFAR-100 | - | - | - | - | **61.71%** | **61.25%** | **61.17%** |
| Plain-56 | CIFAR-100 | - | - | - | - | **61.17%** | **58.30%** | **61.38%** |
| ResNet-44 | CIFAR-100 | 69.41% | 71.29% | 70.69% | 71.11% | **71.28%** | **71.69%** | **71.99%** |
| ResNet-56 | CIFAR-100 | 69.48% | 71.34% | 72.23% | 70.69% | **71.90%** | **72.06%** | **72.28%** |

Our method can be used with BN to further improve the performance. We used CIFAR-10 as a test dataset and tested the performance of the plain CNNs with BN using ReLUs and NG-ReLUs, as shown in table 4. The test accuracy using NG-ReLUs with BN was the highest in all experiments. Please see the appendix for more comparisons about the plain CNNs with BN.

Table 4 Test accuracy comparisons of plain CNNs with BN

| Model depth | ReLU with BN | NG-ReLU | NG-ReLU with BN |
|---|---|---|---|
| 44 | 88.92% | 89.58% | **89.98%** |
| 56 | 87.18% | 88.72% | **89.59%** |

We also used a 20-layer wider ResNet, the structure of which is shown in the appendix, to test performance. The final test accuracy was 95.43%, better than the 1001-layer pre-activation ResNet in [He et al. 2016].

Although NG introduces new training parameters, the extra computational cost is small. For 56-layer plain CNN with BN, NG-ReLUs took 72 seconds on average to run an epoch, whereas ReLUs and PReLUs take 68 and 77 seconds on average. For 56-layer ResNet, NG-ReLUs take 94 seconds on average per epoch, whereas ReLUs and PReLUs took 87 seconds and 100 seconds on average per epoch.

**Exploring Very Deep Model**

We explored the critical depths of the plain CNNs with NGs. The model depth was gradually increased by integral multiples of 12 layers, which is the same as [he et al. 2016]. The test dataset was CIFAR-10. Table 5 shows the results of three weight initializations. The critical depths for our method were much greater.

Table 5 Critical depth of different model

| Model | Critical depth |
|---|---|
| SELU-MSRA | <20 |
| SELU-Xavier | 44 |
| SELU-Orthogonal | 68 |
| ReLU-MSRA | 20 |
| ReLU-Xavier | <20 |
| ReLU-Orthogonal | <20 |
| LReLU-MSRA | 20 |
| LReLU-Xavier | <20 |
| LReLU-Orthogonal | <20 |
| PReLU-MSRA | 20 |
| PReLU-Xavier | 20 |
| PReLU-Orthogonal | 20 |
| NG-ReLU-MSRA | 80 |
| NG-ReLU-Xavier | 116 |
| NG-ReLU-Orthogonal | 152 |
| NG-LReLU-MSRA | 68 |
| NG-LReLU-Xavier | 104 |
| NG-LReLU-Orthogonal | 128 |
| NG-PReLU-MSRA | 128 |
| NG-PReLU-Xavier | 128 |
| NG-PReLU-Orthogonal | 152 |

## Conclusion

In this paper, we proposed the nonlinearity generation method that begins training with a relatively low-capacity model and gradually improves model capacity. The proposed training method modifies a family of activation functions by introducing a trainable parameter $t$ to make the activation functions linear symmetric for the inputs, which makes the model with relatively low capacity and easy to optimize at the beginning. Nonlinearity is then introduced automatically during the training procedure to endow the model with greater capacity. The introduced parameters can be easily incorporated into training. The proposed method can be considered an unusual form of regularization during the training procedure: The parameters in the early stages are obtained by training a relatively low-capacity model for only a few iterations, and are reused for the initialization of a higher-capacity model. We derived the upper and lower bounds of the variance for weight variation and showed that the symmetric structure of NGs helps stabilize training.

Experiments showed that the proposed method speeds up the convergence of training, allows for less careful initialization, improves or at least maintains the performance of CNNs at negligible extra computational cost and can be used with BN to further improve the performance. Finally, we can train a very deep model with the proposed method easily.

# Appendix

## 1. Proof of Theorem 1

Proof of theorem 1: Recall the weight update formula for layer $l$,

$$\Delta w_{i,j}^l = -\eta \frac{\partial \varepsilon}{\partial s_i^l} z_j^l.$$

Where $i$, $j$ denote the $i$ th and $j$ th nodes from layer $l$ and layer $l-1$ respectively. $s_i^l$ denotes $\sum_j^n w_{i,j}^l z_j^l$, and $z_j^l = f(s_j^{l-1})$. $f(\cdot)$ is the activation function. $\Delta w_{i,j}^l$ is the weight variation of $w_{i,j}^l$. Then the variance of the weight variation is:

$$Var(\Delta w^l) = \frac{\eta^2}{mn} \sum_i^m \sum_j^n (\frac{\partial \varepsilon}{\partial s_i^l} z_j^l - \frac{1}{mn}\sum_t^m \sum_k^n (\frac{\partial \varepsilon}{\partial s_t^l} z_k^l))^2$$

where $m$ and $n$ are the number of the nodes in layer $l+1$ and layer $l$ respectively. Using the following inequality

$$\sum_i^n a_i^2 \geq \frac{1}{n}(\sum_i^n a_i)^2$$

We have

$$Var(\Delta w^l) \geq \frac{\eta^2}{mn} \sum_j^n (\sum_i^m \frac{\partial \varepsilon}{\partial s_i^l} z_j^l - \frac{1}{n}\sum_t^m \sum_k^n (\frac{\partial \varepsilon}{\partial s_t^l} z_k^l))^2$$

$$= \frac{\eta^2}{m^2 n} \sum_j^n (z_j^l - \bar{z}^l)^2 (\sum_i^m \frac{\partial \varepsilon}{\partial s_i^l})^2$$

$$= \eta^2 Var(z^l) E^2(\frac{\partial \varepsilon}{\partial S^l})$$

Where $\bar{z}^l = \frac{1}{n}\sum_j^n z_j^l$. On the other hand, by denoting $\overline{\frac{\partial \varepsilon}{\partial s^l}}$ as $\frac{1}{m}\sum_i^m \frac{\partial \varepsilon}{\partial s_i^l}$, we have

$$Var(\Delta w^l) = \frac{\eta^2}{mn}\sum_i^m\sum_j^n (\frac{\partial \varepsilon}{\partial s_i^l}(z_j^l - \bar{z}^l + \bar{z}^l) - \bar{z}^l \overline{\frac{\partial \varepsilon}{\partial s^l}})^2$$

$$= \frac{\eta^2}{mn}\sum_i^m\sum_j^n (\bar{z}^l(\frac{\partial \varepsilon}{\partial s_i^l} - \overline{\frac{\partial \varepsilon}{\partial s^l}}) + (z_j^l - \bar{z}^l)\frac{\partial \varepsilon}{\partial s_i^l})^2$$

$$\leq 2\eta^2(\bar{z}^l)^2 \frac{1}{m}\sum_i^m (\frac{\partial \varepsilon}{\partial s_i^l} - \overline{\frac{\partial \varepsilon}{\partial s^l}})^2 + 2\eta^2 \frac{1}{n}\sum_j^n (z_j^l - \bar{z}^l)^2 \frac{1}{m}\sum_i^m (\frac{\partial \varepsilon}{\partial s_i^l})^2$$

$$= 2\eta^2(Var(\frac{\partial \varepsilon}{\partial S^l})E^2(Z^l) + Var(Z^l)Var(\frac{\partial \varepsilon}{\partial S^l}) + Var(Z^l)E^2(\frac{\partial \varepsilon}{\partial S^l})).$$

End the proof.

## 2. Structure details of the test model

Plain CNN: The plain CNNs are 44, 56 layers with the structure shown in table 1.

Table. 1 Plain CNN structure. $[m,n]\times z$ denotes the convolutional layer with kernel size $m$, number of the kernels $n$ and repeating this convolution layer $z$ times.

| 44-layer CNN | 56-layer CNN | Output size |
|---|---|---|
| input | | $32\times 32$ |
| [3,16]× 1 | | $32\times 32$ |
| [3×3,16]× 14 | [3×3,16]× 18 | $32\times 32$ |
| Max Pooling | | |
| [3×3,32]× 14 | [3×3,32]× 18 | $16\times 16$ |
| Max Pooling | | |
| [3×3,64]× 14 | [3×3,64]× 18 | $8\times 8$ |
| Global Average Pooling | | |
| Softmax | | |

ResNet: ResNets are 20, 44, 56 layers with the structure shown in table 2.

Table.2 ResNet structure. $[Block, m, n]\times z$ denotes the identity block, which is same the block used in CIFAR-10 as [He et al. 2016], with kernel size $m$, number of the kernels $n$ and repeating this identity block $z$ times，downsampleing is performed with a stride of 2

| 44-layer ResNet | 56-layer ResNet | 20-layer wider ResNet | Output size |
|---|---|---|---|
| Input | | | $32\times 32$ |
| [3×3,16]× 1 | | [3×3,128]× 1 | $32\times 32$ |
| [Block，3×3, 16]× 7 | [Block，3×3, 16]× 9 | [Block，3×3, 128]× 3 | $32\times 32$ |
| [Block，3×3, 32]× 1, stride 2 <br> [Block，3×3, 32]× 6 | [Block，3×3, 32]× 1, stride 2 <br> [Block，3×3, 32]× 8 | [Block，3×3, 256]× 1, stride 2 <br> [Block，3×3, 256]× 2 | $16\times 16$ |
| [Block，3×3, 64]× 1, stride 2 <br> [Block，3×3, 64]× 6 | [Block，3×3, 64]× 1, stride 2 <br> [Block，3×3, 64]× 8 | [Block，3×3, 512]× 1, stride 2 <br> [Block，3×3, 512]× 2 | $8\times 8$ |
| Global Average pooling | | | |
| Softmax | | | |

## 3. Experiments for plain CNN with batch normalization

### 3.1 Learning Behavior

We tested the 56-layer plain CNNs with batch normalization (BN) and made comparisons on CIFAR10 dataset. We used the same training strategies as the paper. From Fig. 1 we saw that similar to the experiments of the plain CNNs without BN, NGs were insensitive to the weight initializations, while the other models processed relatively different behaviors with different initializations. In addition, the models with NGs converged faster than their counterparts in the most cases. NG also enables larger learning rate (lr), as shown in Fig. 2, the models with NG-ReLUs converge with very large lrs, while the models with ReLUs do not.

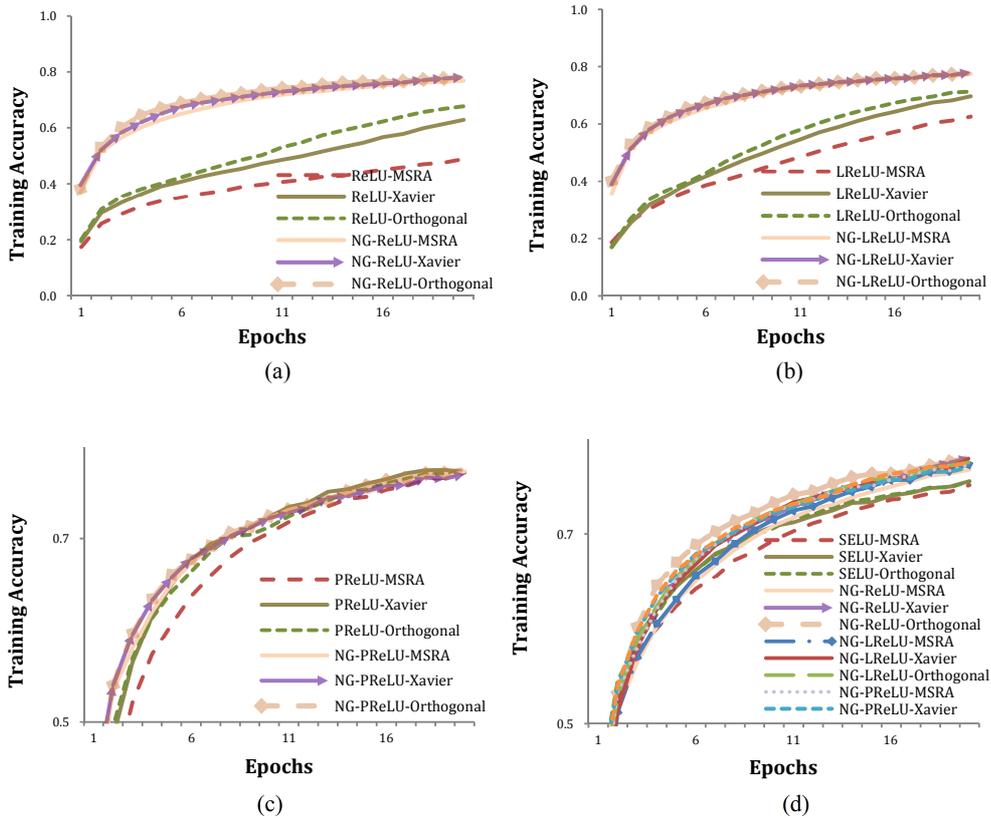

Figure. 1 Learning behavior comparisons with different initializations on CIFAR-10.

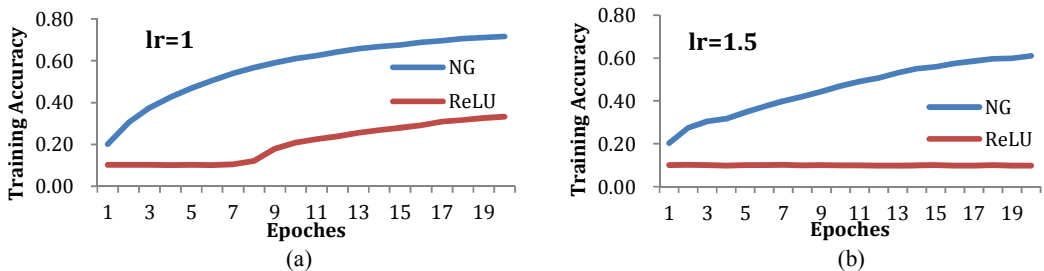

Figure. 2 Learning behaviors of 56-layer plain CNNs with BN using different learning rates. (a), lr = 1; (b) lr = 1.5.

We extracted the weight variance of every layer for the 56-layer plain CNNs and made comparisons. Figure 3 shows the comparisons. Compared with ReLUs, the weight variance using NG-ReLUs is more stable, which supports our analysis in section 3.2 of the paper.

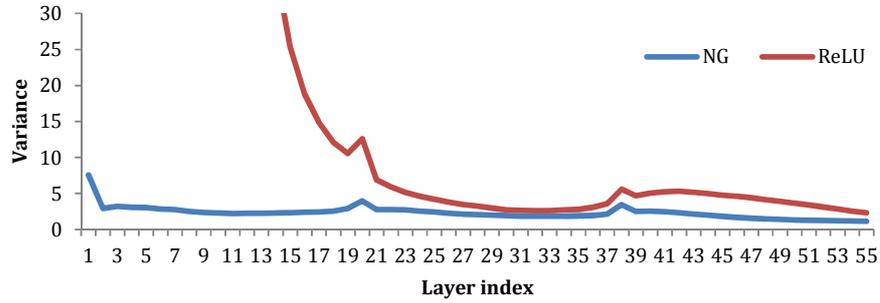

Figure. 3 Weight variance comparison of every layer for the plain CNNs. The test model is 56-layer plain CNN with BN.

### 3.2 Model Performance

We tested and compared 44-layer and 56-layer plain CNNs on CIFAR-10 and SVHN datasets. MSRA initialization was used. Table 3 shows the test accuracy comparisons. Similar to the results of the plain CNNs without BN, NGs perform slightly better or comparable results.

Table 3 Test accuracy comparisons on different datasets.

| Model | Dataset | ReLU | LReLU | PReLU | SELU | NG-ReLU | NG-LReLU | NG-PReLU |
|---|---|---|---|---|---|---|---|---|
| plain-44 CNN | CIFAR-10 | 88.92% | 89.01% | 89.23% | 89.09% | **89.48%** | **89.06%** | **89.70%** |
| plain-56 CNN | CIFAR-10 | 87.18% | 86.02% | 89.13% | 88.30% | **89.40%** | **88.90%** | **89.22%** |
| plain-44 CNN | SVHN | 95.99% | 95.80% | 96.30% | 95.39% | **96.07%** | **95.91%** | **96.31%** |
| plain-56 CNN | SVHN | 95.79% | 95.50% | 96.14% | 95.59% | **96.12%** | **95.84%** | 96.10% |

### 3.3 Exploring Very Deep Model

Finally, we explored the critical depths of the plain CNNs with BN. The test dataset was CIFAR-10. We could not test the learning behaviors of the plain CNNs with more than 248 layers because of GPU memory constraints. So the largest depth was 248 layers in this subsection. Table 4 shows the results of ReLU and NG-ReLU with three weight initializations. The critical depths NG-ReLU are much greater than ReLU.

Table 4 Critical depth of different model

| Model | Critical depth |
|---|---|
| SELU-MSRA | 224 |
| SELU-Xavier | 248 |
| SELU-Orthogonal | 236 |
| ReLU-MSRA | 104 |
| ReLU-Xavier | 80 |
| ReLU-Orthogonal | 116 |
| LReLU-MSRA | 80 |
| LReLU-Xavier | 80 |
| LReLU-Orthogonal | 116 |
| PReLU-MSRA | 200 |
| PReLU-Xavier | 188 |
| PReLU-Orthogonal | 224 |
| NG-ReLU-MSRA | 248 |
| NG-ReLU-Xavier | 248 |
| NG-ReLU-Orthogonal | 236 |
| NG-LReLU-MSRA | 248 |
| NG-LReLU-Xavier | 248 |
| NG-LReLU-Orthogonal | 248 |
| NG-PReLU-MSRA | 248 |
| NG-PReLU-Xavier | 248 |
| NG-PReLU-Orthogonal | 248 |